%% file: colchunk.tex
\documentclass[sigconf]{acmart}
\AtBeginDocument{%
  }

\setcopyright{acmlicensed}
\copyrightyear{2018}
\acmYear{2018}
\acmDOI{XXXXXXX.XXXXXXX}
\acmConference[Conference acronym 'XX]{Make sure to enter the correct
  conference title from your rights confirmation email}{June 03--05,
  2018}{Woodstock, NY}
\acmISBN{978-1-4503-XXXX-X/2018/06}

\usepackage{hyperref}
\usepackage{url}
\usepackage{amsthm}
\usepackage{hyperref}

\usepackage[most]{tcolorbox}
\usepackage{wrapfig}
\usepackage{graphicx,float}
\usepackage{subfigure}
\usepackage{threeparttable}
\usepackage[ruled,vlined]{algorithm2e}
\usepackage{colortbl}
\usepackage{color}
\usepackage{multirow}
\usepackage{tabularx}
\usepackage{float}
\usepackage{graphicx}
\usepackage{booktabs}
\usepackage{arydshln}
\usepackage{enumitem}
\usepackage{wrapfig}
\usepackage{caption}
\usepackage{graphicx}
\usepackage{makecell}
\usepackage{float} 
\usepackage{mathtools}
\usepackage{subfigure}
\usepackage{bbding}
\usepackage{makecell}
\usepackage{tabularx}

\usepackage{amssymb,mathrsfs,amsmath}
\usepackage{pifont}
\usepackage{bbm}
\usepackage{amsmath}
\usepackage{algpseudocode}

\usepackage{listings}

\definecolor{bittersweet}{rgb}{1.0, 0.44, 0.37}
\definecolor{mygreen}{rgb}{0.29, 0.7, 0.48}
\usepackage{pifont}
\definecolor{demphcolor}{RGB}{144,144,144}

\definecolor{mygray}{gray}{0.4}
\definecolor{autopurple}{HTML}{7030A0}
\definecolor{dyna_yellow}{HTML}{BF9000}
\definecolor{adaptive_blue}{HTML}{0070C0}
\definecolor{darksalmon}{rgb}{0.91, 0.59, 0.48}
\definecolor{emerald}{rgb}{0.31, 0.78, 0.47}
\definecolor{green(pigment)}{rgb}{0.0, 0.65, 0.31}
\definecolor{amaranth}{rgb}{0.9, 0.17, 0.31}
\definecolor{iris}{rgb}{0.35, 0.31, 0.81}
\definecolor{uu}{rgb}{0.95, 0.51, 0.51}
\definecolor{spirodiscoball}{rgb}{0.06, 0.75, 0.99}
\usepackage{svg}

\usepackage{cleveref}
\SetKwInOut{Input}{Input}\SetKwInOut{Output}{Output}

\SetCommentSty{mycommfont}
\SetKwComment{Comment}{$\triangleright$\ }{}

\usepackage{xspace}
\newcommand{\ourmethod}{{\fontfamily{lmtt}\selectfont \textbf{ColChunk}}\xspace}

\definecolor{ada_blue}{rgb}{0,205,205}
\definecolor{glt_red}{rgb}{109,205,255}
\definecolor{MorandiBlue}{RGB}{118,134,146}

\definecolor{demphcolor}{RGB}{144,144,144}
\definecolor{mygray}{gray}{0.4}
\definecolor{autopurple}{HTML}{7030A0}
\definecolor{dyna_yellow}{HTML}{BF9000}
\definecolor{adaptive_blue}{HTML}{0070C0}
\definecolor{darkgrey}{RGB}{120,120,120}
\definecolor{mygrey}{RGB}{200,200,200}

\usepackage[font=small,labelfont=bf]{caption}  
\usepackage{makecell}
\usepackage{tabulary}

\definecolor{myblue}{HTML}{00CDCD}
\definecolor{champagne}{rgb}{0.74, 0.83, 0.9}
\definecolor{champagne}{rgb}{0.97, 0.91, 0.81}

\hypersetup{
    colorlinks=true,
    linkcolor=red,
    citecolor=cyan,
    filecolor=magenta,      
    urlcolor=magenta,
    }

\begin{document}

\title{Visual Late Chunking: An Empirical Study of Contextual Chunking for Efficient Visual Document Retrieval}

\author{\textbf{Yibo Yan}$^{1,2,3}$, 
    \textbf{Mingdong Ou}$^{2}$, 
    \textbf{Yi Cao}$^{2}$, 
    \textbf{Jiahao Huo}$^{1,2}$,
    \textbf{Xin Zou}$^{1,3}$, 
    \textbf{Shuliang Liu}$^{1,3}$, \\
    \textbf{James Kwok}$^{3}$, 
    \textbf{Xuming Hu}$^{1,3,*}$\\
    $^1$Hong Kong University of Science and Technology (Guangzhou), \\
    $^2$Alibaba Cloud Computing, 
    $^3$Hong Kong University of Science and Technology
}
\renewcommand{\shortauthors}{Trovato et al.}

\begin{abstract}
Multi-vector models dominate Visual Document Retrieval (VDR) due to their fine-grained matching capabilities, but their high storage and computational costs present a major barrier to practical deployment. In this paper, we propose \textbf{\ourmethod{}}, a plug-and-play framework that introduces \textit{multimodal late chunking} to construct efficient, contextualized multi-vectors. Unlike existing pruning or fixed-token approaches, \ourmethod{} employs hierarchical clustering on patch-level embeddings, fused with a 2D position prior to ensure spatial-semantic coherence. This adaptive grouping allows for a content-aware representation that preserves global context while drastically reducing the vector count. Evaluations across 24 VDR datasets demonstrate \ourmethod{} achieves over a 90\% reduction in storage requirements while simultaneously delivering a 9-point average improvement in nDCG@5 across representative single-vector models. \ourmethod provides a practical solution for balancing retrieval accuracy and efficiency in visual document systems.
\end{abstract}





\maketitle

\section{Introduction}
\label{sec:introduction}
\input{sections/introduction}

\section{Related Work}
\label{sec:related_work}
\input{sections/related_work}

\section{Methodology}
\label{sec:method}
\input{sections/method}

\section{Experiment}
\label{sec:experiment}
\input{sections/experiment}

\section{Conclusion}
\label{sec:conclusion}
\input{sections/conclusion}

\clearpage
\bibliographystyle{ACM-Reference-Format}
\bibliography{software,colchunk}


\end{document}

%% file: sections/introduction.tex
Visual Document Retrieval (VDR) has emerged as a critical task in information retrieval, aimed at identifying relevant documents from massive corpora based on both textual content and visual layout \cite{ding2025survey,gao2025scaling,yan2026unlocking}. 
Currently, the state-of-the-art paradigm is dominated by multi-vector architectures, epitomized by ColPali \cite{faysse2024colpali}. 
These models represent each document page as a collection of patch-level embeddings, a strategy that perfectly aligns with the nature of visual documents where semantic information is densely packed and spatially distributed. 
By allowing ``late interaction'' between query terms and document patches, multi-vector models preserve fine-grained details that are often lost in coarse-grained single-vector representations \cite{khattab2020colbert,santhanam2022colbertv2,qian2022multi,liu2023understanding}.

Despite their superior retrieval accuracy, multi-vector VDR models face a significant bottleneck: \textit{storage efficiency} \cite{lee2023rethinking,park2025scv,scheerer2025warp,shrestha2024espn}. 
Storing thousands of embeddings per page leads to prohibitive memory costs in large-scale applications. 
To mitigate this, several optimization routes have been explored. 
\ding{182} \textit{Merging-based} schemes, such as Light-ColPali \cite{ma2025towards}, group similar tokens to reduce count \cite{clavie2024reducing,cha2026reinpool,macavaney2025efficient}, yet often suffer from unstable performance due to over-simplified aggregation. 
\ding{183} \textit{Pruning-based} methods, like DocPruner \cite{yan2025docpruner}, discard ``redundant'' tokens \cite{lassance2022learned,he2025token,veneroso2025crisp,yan2026sculpting}, but their performance degrades sharply under high compression ratios as critical context is lost. 
\ding{184} More recently, approaches like MetaEmbed \cite{xiao2025metaembed} and CausalEmbed \cite{huo2026causalembed} \textit{introduce additional trainable ``summary'' tokens}. 
However, these often require expensive re-training and lack the flexibility to adapt to varying document complexities, as they rely on a fixed number of learned latent representations.

A potential solution lies in the concept of ``\textbf{late chunking},'' proposed by \citet{gunther2024late} and recently popularized in text embedding research \cite{merola2025reconstructing,conti2025context,wang2025jinarerankerv3}. 
In the text domain, late chunking leverages long-context models to generate token-level embeddings first, applying pooling only at the final stage to ensure each chunk embedding is richly contextualized. 
Inspired by this, we argue that a similar ``multimodal late chunking'' is essential for VDR. 
However, while text chunking is inherently sequential, visual documents are non-linear and spatial. 
Our motivation is to move beyond rigid or learned tokens and instead construct contextualized multi-vectors that are content-aware, allowing the number of chunks to be dynamically set to balance efficiency and semantic richness.

Therefore, we propose \ourmethod{}, a novel paradigm for constructing contextualized multi-vectors specifically designed for visual documents. \ourmethod{} overcomes the limitations of previous methods by performing hierarchical clustering on the semantic embeddings of the final transformer layer. By grouping tokens with similar semantic signatures into adaptive chunks and applying pooling, we generate a condensed set of multi-vectors for \texttt{MaxSim}-based retrieval. Crucially, to integrate the document's spatial structure, we incorporate a \textit{position prior}. By fusing 2D positional encodings with semantic embeddings during the clustering process, \ourmethod{} ensures that the resulting chunks are both semantically coherent and spatially contiguous. This training-free approach retains global context while enabling a content-sensitive representation.

We conduct extensive experiments across 24 VDR datasets, adapting \ourmethod{} to five representative single-vector models. The results are compelling: by using only 40 multi-vectors per page, a reduction of over 90\% compared to original multi-vector counts (\textit{e.g.,} avg 768 in ColQwen2.5 \cite{faysse2024colpali}), we achieve an average improvement of 9 points in nDCG@5 compared to the base single-vector models. This demonstrates that \ourmethod{} serves as a flexible ``plug-and-play'' enhancement, bridging the gap between efficient single-vector models and high-performance multi-vector mechanisms while deeply respecting the characteristics of visual documents.

Our contributions are summarized as follows:
\begin{itemize}[leftmargin=*]
    \item[\ding{182}] \textbf{Novel Paradigm:} We are the first to migrate the concept of late chunking to the visual document domain, addressing the critical storage bottleneck of current multi-vector VDR models.
    \item[\ding{183}] \textbf{Training-free Flexibility:} We propose a clustering-based framework with a spatial-semantic position prior that can be seamlessly applied to various models without additional training.
    \item[\ding{184}] \textbf{Superior Trade-off:} The extensive experiment demonstrate \ourmethod{} achieves a superior trade-off between performance and efficiency, providing a foundation for real-world deployment.
\end{itemize}

%% file: sections/related_work.tex
VDR is essential for accessing information in visually rich documents where semantic meaning is derived from both textual content and spatial layout \cite{yan2026unlocking,yan2026beyond}. 
Traditional pipelines rely on OCR to extract text, but they often \textit{struggle to preserve structural integrity and fail on non-textual elements like tables and charts} \cite{zhang2025ocr,most2025lost}. 
While the era of Large Vision-Language Models (LVLMs) has introduced end-to-end single-vector models to bypass OCR (\textit{e.g.,} DSE \cite{ma2024dse}, GME \cite{zhang2024gme}, UniSE \cite{liu2025any}), these models \textit{suffer from significant information loss by compressing complex, high-resolution pages into a single coarse-grained representation}. 
Consequently, multi-vector architectures, pioneered by ColPali \cite{faysse2024colpali}, have redefined the SOTA through late interaction \cite{khattab2020colbert}, with subsequent research focusing on enhancing performance via model architecture (\textit{e.g.,} ModernVBERT \cite{teiletche2025modernvbert}), data synthesis (\textit{e.g.,} Nemotron ColEmbed V2 \cite{moreira2026nemotron}), and training objectives (\textit{e.g.,} jina-embeddings-v4 \cite{gunther2025jina}). 
Despite their superior accuracy, these models face a \textit{severe efficiency bottleneck} due to the prohibitive storage of maintaining thousands of patch-level embeddings per page. 
To mitigate this footprint, current research generally follows three paradigms: pruning redundant embeddings (\textit{e.g.,} DocPruner \cite{yan2025docpruner}), merging similar tokens via pooling or clustering (\textit{e.g.,} Light-ColPali \cite{ma2025towards}), or introducing learnable summary tokens (\textit{e.g.,} MetaEmbed \cite{xiao2025metaembed}). 
In this context, \ourmethod addresses these limitations by migrating the concept of late chunking to the visual domain through spatial-semantic aware clustering.

%% file: sections/method.tex
\subsection{Task Formulation}
\label{sec:task_setting}

The task of VDR is to identify the most relevant document page $d$ from a corpus $\mathcal{C}$ for a given textual query $q$. We follow the multi-vector retrieval paradigm, utilizing an LVLM-based encoder $\Phi(\cdot)$ to map inputs into a continuous embedding space. 

Formally, a query $q$ is encoded into a set of $N_q$ token-level embeddings $\mathbf{Q} = \{\mathbf{q}_i\}_{i=1}^{N_q}$, where each $\mathbf{q}_i \in \mathbb{R}^D$. Similarly, a document page $d$, processed as an image, is represented by a set of $N_v$ patch-level contextual embeddings $\mathbf{V} = \{\mathbf{v}_j\}_{j=1}^{N_v}$, where $\mathbf{v}_j \in \mathbb{R}^D$. The relevance score $S(q, d)$ is computed using the late interaction mechanism:
\vspace{-2mm}
\begin{equation}
    S(q, d) = \sum_{i=1}^{N_q} \max_{j=1}^{N_v} \frac{\mathbf{q}_i^\top \mathbf{v}_j}{\|\mathbf{q}_i\| \|\mathbf{v}_j\|}.
\end{equation}
We aim to construct a condensed multi-vector set $\mathbf{D} = \{\mathbf{d}_k\}_{k=1}^{K}$ with size $K \ll N_v$ (\textit{e.g.,} $K=40$), minimizing storage overhead while preserving retrieval precision through spatial-semantic coherence.

\begin{algorithm}[!ht]
\caption{\ourmethod Compression Workflow}\label{alg:colchunk}
\Input{Document image $d$, Pre-trained LVLM encoder $\Phi(\cdot)$, Target number of chunks $K$, Position prior weight $\omega$, Positional encoding function $f_{pos}(\cdot)$}
\Output{Compressed multi-vector set $\mathbf{D} = \{\mathbf{d}_k\}_{k=1}^K$}

\textcolor{blue}{\tcc{Phase 1: Contextual Feature Extraction \& Fusion}}
Extract patch-level contextual embeddings from the last hidden layer of $\Phi$: \\
$\mathbf{V} = \{\mathbf{v}_1, \mathbf{v}_2, \dots, \mathbf{v}_{N_v}\} \leftarrow \Phi(d)$, where $\mathbf{v}_j \in \mathbb{R}^D$\;

\For{each patch $j$ in $\{1, 2, \dots, N_v\}$}{
    Obtain normalized 2D coordinates $P_j = (x_j, y_j)$\;
    Generate positional embedding $\mathbf{p}_j \leftarrow f_{pos}(P_j)$\;
    Compute spatial-semantic fused feature: $\mathbf{z}_j \leftarrow (1 - \omega) \cdot \mathbf{v}_j + \omega \cdot \mathbf{p}_j$ \tcp*{$\omega$ balances layout and semantics}
}
Gather fused feature set $\mathbf{Z} \leftarrow \{\mathbf{z}_1, \mathbf{z}_2, \dots, \mathbf{z}_{N_v}\}$\;

\BlankLine
\textcolor{blue}{\tcc{Phase 2: Hierarchical Contextual Chunking}}
Initialize Hierarchical Agglomerative Clustering on $\mathbf{Z}$\;
\While{number of clusters $> K$}{
    Find and merge two clusters that minimize Ward's linkage distance based on $\mathbf{Z}$\;
}
Obtain cluster assignments $\{\mathcal{C}_1, \mathcal{C}_2, \dots, \mathcal{C}_K\}$ \tcp*{Each $\mathcal{C}_k$ contains indices of patches in the $k$-th chunk}

\BlankLine
\textcolor{blue}{\tcc{Phase 3: Representative Multi-Vector Generation}}
\For{each cluster $\mathcal{C}_k$ in $\{\mathcal{C}_1, \dots, \mathcal{C}_K\}$}{
    Get cluster size $n_k \leftarrow |\mathcal{C}_k|$\;
    Aggregate original semantic embeddings: $\bar{\mathbf{d}}_k \leftarrow \frac{1}{n_k} \sum_{j \in \mathcal{C}_k} \mathbf{v}_j$ \tcp*{Centroid of the $k$-th chunk}
    Apply $L_2$ normalization: $\mathbf{d}_k \leftarrow \bar{\mathbf{d}}_k / \|\bar{\mathbf{d}}_k\|_2$\;
}
\Return $\mathbf{D} \leftarrow \{\mathbf{d}_1, \mathbf{d}_2, \dots, \mathbf{d}_K\}$ \tcp*{Final multi-vector set}

\end{algorithm}

\subsection{The \ourmethod\ Framework}
\label{sec:our_framework}

\ourmethod\ is an offline compression framework designed to transform patch-level embeddings into a streamlined yet semantically potent set of chunk embeddings. Given the set of $N_v$ contextualized patch embeddings $\mathbf{V} = \{\mathbf{v}_j\}_{j=1}^{N_v}$ where $\mathbf{v}_j \in \mathbb{R}^D$, extracted from an LVLM encoder, \ourmethod\ constructs a multi-vector representation via three sequential stages as shown in Algorithm \ref{alg:colchunk}.

\subsubsection{Spatial-Semantic Feature Integration}
The semantics of visual documents are intrinsically tied to their spatial layout. To ensure that the subsequent clustering process is layout-aware, we map the 2D coordinates of each patch into the embedding space. For each patch $j \in \{1, \dots, N_v\}$, its position in the normalized coordinate system is defined as $P_j = (x_j, y_j)$. We apply a 2D sinusoidal encoding function $f_{pos}: \mathbb{R}^2 \to \mathbb{R}^D$ to derive the positional embedding $\mathbf{p}_j \in \mathbb{R}^D$. The enhanced feature $\mathbf{z}_j \in \mathbb{R}^D$ is then formed by fusing the semantic embedding $\mathbf{v}_j$ with the positional embedding $\mathbf{p}_j$ via a weighting factor $\omega \in [0, 1]$: $\mathbf{z}_j = (1 - \omega) \cdot \mathbf{v}_j + \omega \cdot \mathbf{p}_j,$
where $\omega$ serves as a hyperparameter to balance the influence of semantic similarity and spatial proximity during clustering.

\subsubsection{Hierarchical Contextual Chunking}
\ourmethod\ implements \textbf{content-aware} partitioning through Hierarchical Agglomerative Clustering (HAC). we define an assignment function $\mathcal{A}: \{1, \dots, N_v\} \to \{1, \dots, K\}$ that partitions the $N_v$ patches into $K$ disjoint chunks $\{\mathcal{C}_1, \mathcal{C}_2, \dots, \mathcal{C}_K\}$, where $\mathcal{C}_k = \{j \mid \mathcal{A}(j) = k\}$.

A key characteristic of \ourmethod\ is that the number of tokens assigned to each chunk is adaptive, denoted as $n_k = |\mathcal{C}_k|$, satisfying $\sum_{k=1}^K n_k = N_v$. This non-uniformity allows the framework to dynamically adapt to document elements of varying scales. For example, a small $n_k$ (\textit{e.g.,} $n_k \approx 4$) may represent sparse elements such as a page number or an isolated icon; whereas a large $n_k$ (\textit{e.g.,} $n_k \approx 120$) typically corresponds to a dense text paragraph or a complex graphical region.
By utilizing Ward's linkage criterion to minimize intra-cluster variance, \ourmethod\ ensures that each chunk is both spatially contiguous and semantically coherent.

\subsubsection{Representative Multi-Vector Generation}
Following the cluster assignment, we generate the final document multi-vector set $\mathbf{D} \in \mathbb{R}^{K \times D}$ by aggregating the original semantic embeddings $\mathbf{v}_j$ (excluding the positional prior $\mathbf{p}_j$ to avoid geometric bias in retrieval). For the $k$-th chunk, the representative embedding $\mathbf{d}_k \in \mathbb{R}^D$ is calculated as: $\mathbf{d}_k = \text{Norm} \left( \frac{1}{n_k} \sum_{j \in \mathcal{C}_k} \mathbf{v}_j \right),$
where $\text{Norm}(\cdot)$ denotes $L_2$ normalization operator. 
The resulting set $\mathbf{D} = \{\mathbf{d}_k\}_{k=1}^K$ serves as final efficient index. Since each $\mathbf{d}_k$ is derived from late hidden states that have already integrated global context, the representation preserves fine-grained details even at a high compression\%.

\begin{figure}[!t]
  \centering
  \includegraphics[width=0.7\linewidth]{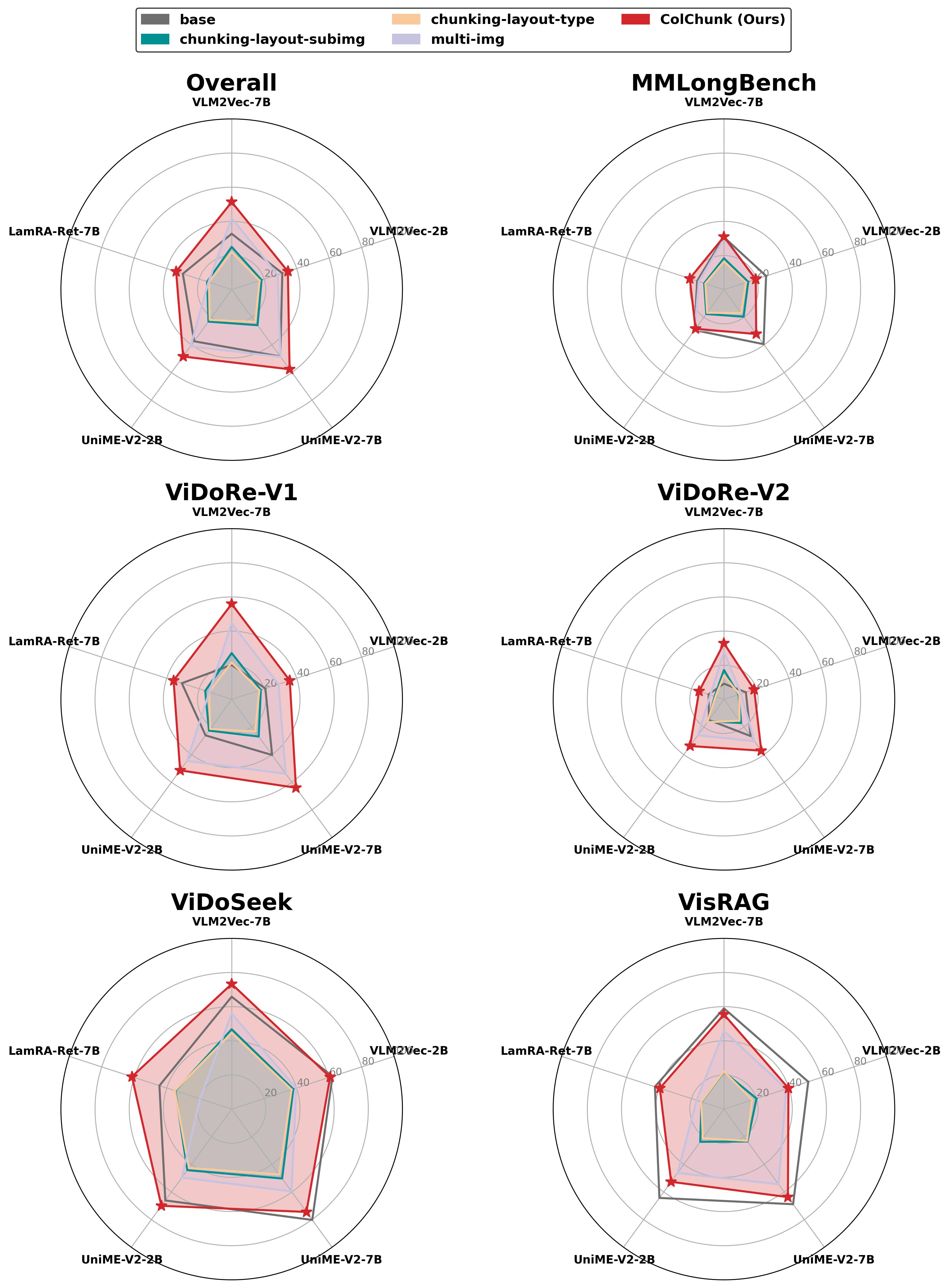}
  \vspace{-2mm}
  \caption{The performance comparison of \ourmethod on five VDR benchmarks across five single-vector retrieval models.}
   \label{fig:main_result_visualization}
   \vspace{-6mm}
\end{figure}

\subsection{Theoretical Foundation}

The efficacy of \ourmethod~can be understood through the Information Bottleneck (IB) principle \cite{tishby2000information}. Given a document $d$ and query $q$, the information flow forms a Markov chain $q \rightarrow d \rightarrow \mathbf{E}_{\text{patch}} \rightarrow \mathbf{C}$. The Data Processing Inequality implies $I(q; \mathbf{C}) \le I(q; \mathbf{E}_{\text{patch}})$, posing the challenge of minimizing information loss during compression. The IB objective is to find a compressed representation $\mathbf{C}$ by solving $\min_{\mathbf{C}} [I(\mathbf{C}; \mathbf{E}_{\text{patch}}) - \beta I(\mathbf{C}; q)]$. Since $q$ is unknown at indexing time, we use the document's spatial-semantic structure $d_{\text{struct}}$ as a tractable proxy for $q$, reformulating the objective as $\min_{\mathbf{C}} [I(\mathbf{C}; \mathbf{E}_{\text{patch}}) - \beta I(\mathbf{C}; d_{\text{struct}})]$. \ourmethod~approximates this optimization in a training-free manner. The clustering and pooling stage, $g: \mathbf{E}_{\text{patch}} \mapsto \mathbf{C}$, acts as a quantizer that minimizes $I(\mathbf{C}; \mathbf{E}_{\text{patch}})$ by grouping $N$ patches into $k$ centroids. The spatial-semantic fusion stage maximizes the ``relevance'' term $I(\mathbf{C}; d_{\text{struct}})$ by ensuring that the clustering operates on features $\mathbf{F}$ that encode both semantics and spatial layout. This forces the resulting chunks $\mathbf{C}$ to preserve the document's salient structural information.

%% file: sections/experiment.tex
\subsection{Experimental Setup}

\textbf{Benchmarks \& Evaluation.} 
We evaluate \ourmethod across five prominent VDR benchmarks encompassing a total of 24 datasets: ViDoRe-V1 \cite{faysse2024colpali}, ViDoRe-V2 \cite{mace2025vidorev2}, VisRAG \cite{yu2024visrag}, ViDoSeek \cite{wang2025vidorag}, and MMLongBench \cite{ma2024mmlongbench}. To demonstrate the robustness of \ourmethod, we conduct comprehensive verification on five representative single-vector models: VLM2Vec-2B/7B \cite{jiang2024vlm2vec}, LamRA-Ret \cite{liu2025lamra}, and UniME-V2-2B/7B \cite{gu2025unimev2}. We employ nDCG@5 as our primary metric.

\begin{figure}[!t]
  \centering
  \includegraphics[width=0.65\linewidth]{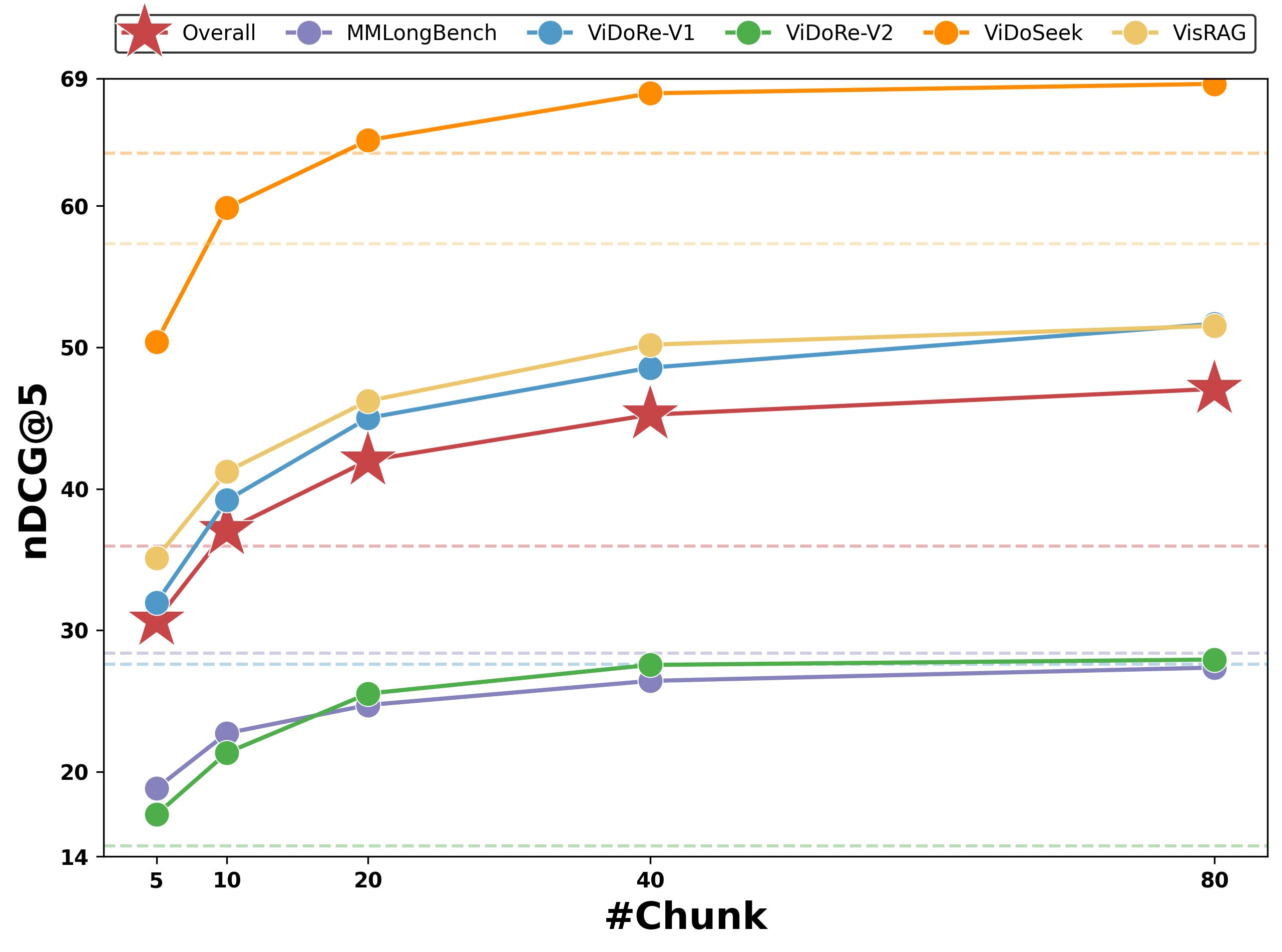}
  \vspace{-5mm}
  \caption{The comparison of the \textit{average} performance of \ourmethod across different chunk size. The dash lines refer to the base results.}
   \label{fig:chunk_size_scaling}
   \vspace{-6mm}
\end{figure}

\begin{figure*}[!t]
  \centering
  \begin{minipage}{0.29\linewidth}
    \centering
    \includegraphics[width=\linewidth]{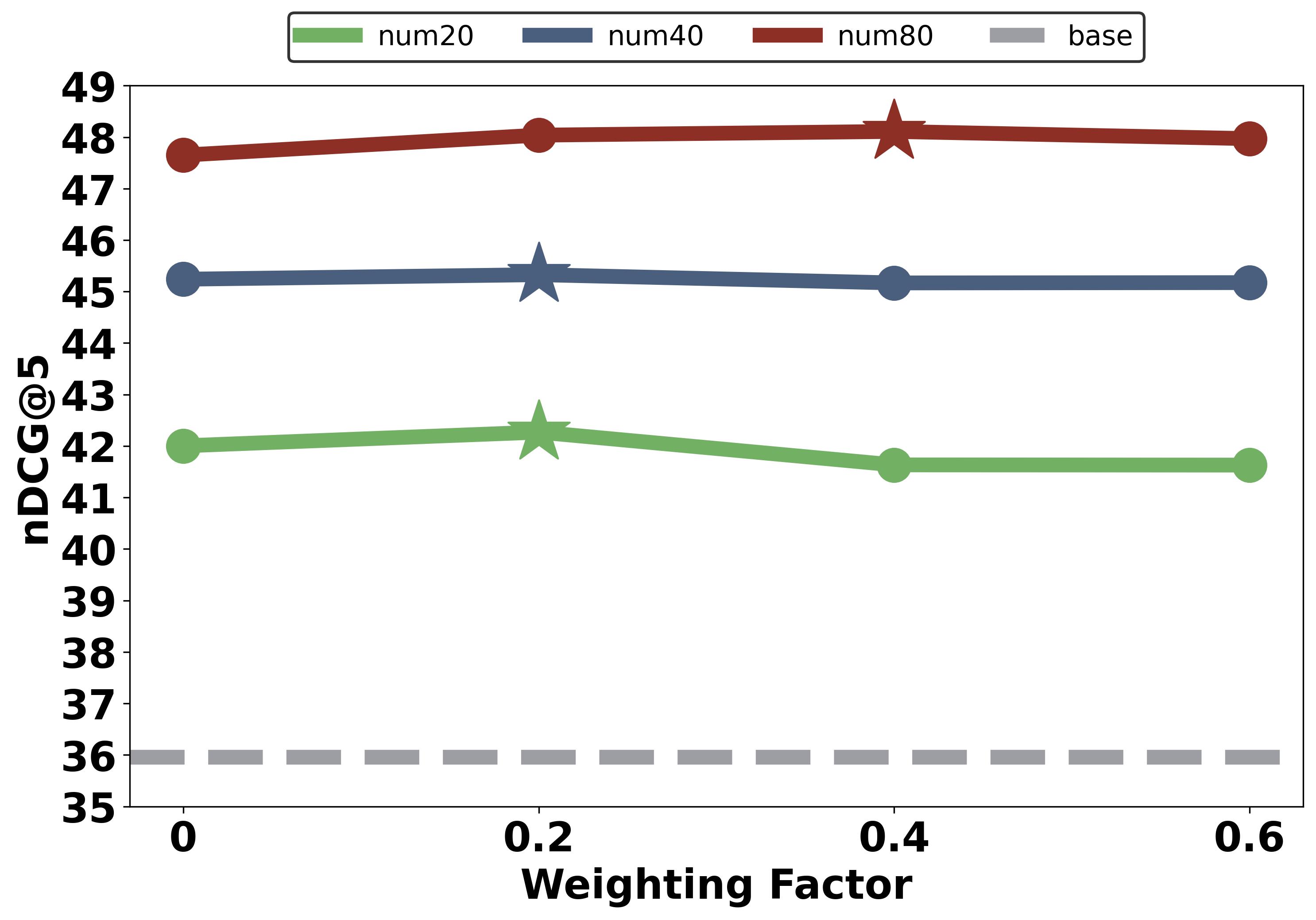}
  \end{minipage}
  \hfill
  \begin{minipage}{0.34\linewidth}
    \centering
    \includegraphics[width=\linewidth]{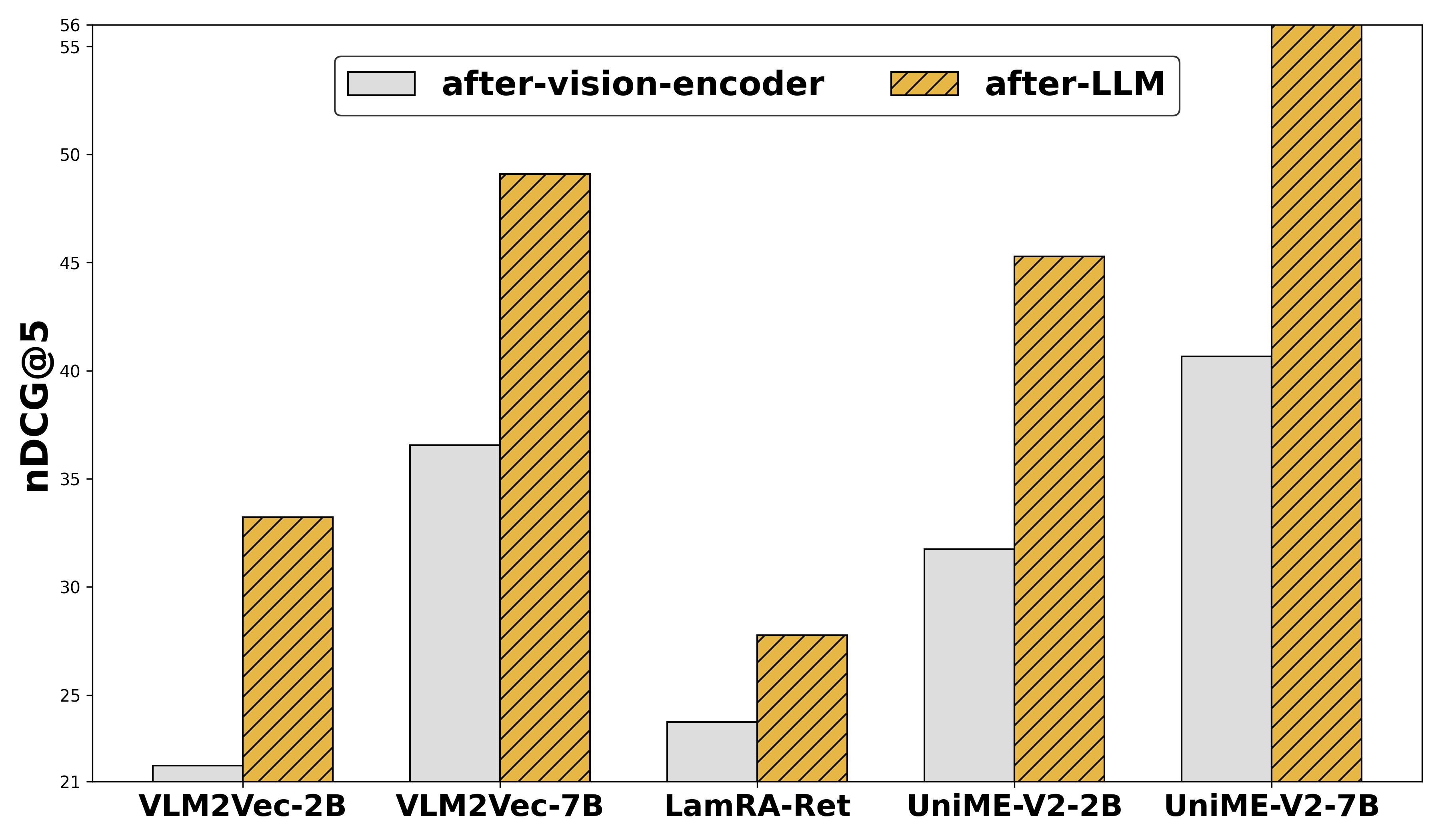}
  \end{minipage}
  \hfill
  \begin{minipage}{0.34\linewidth}
    \centering
    \includegraphics[width=\linewidth]{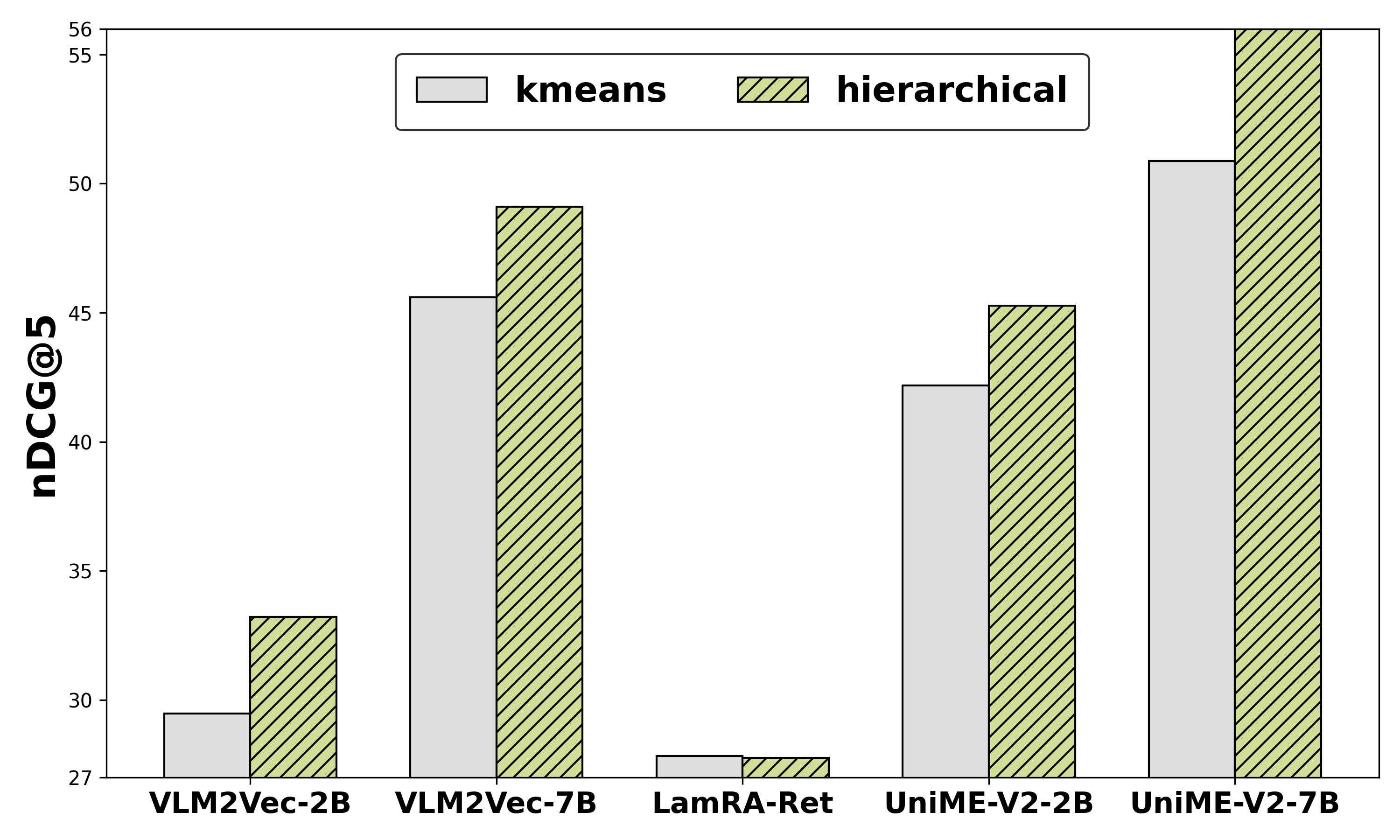}
  \end{minipage}

  \vspace{-2mm}
  \caption{The comparison of the \textit{overall} performance of \ourmethod: 
  \textbf{\textit{(Left)}}: with different weighting factors (different color indicates different chunk size, dash lines refer to average base results, and stars refer to the best results); 
  \textbf{\textit{(Middle)}}: with different chunking positions (after vision encoder vs. after LLM backbone) across five models; 
  \textbf{\textit{(Right)}}: with different clustering methods (k-means vs. hierarchical) across five models.}
  \label{fig:hyperparameters_comparison}
  \vspace{-4mm}
\end{figure*}

\textbf{Baselines.} 
We compare \ourmethod against three categories of baselines: (I) \textbf{Base}, representing the original single-vector models without multi-vector expansion; (II) \textbf{Multi-img}, where all sub-images extracted via parsing are simultaneously fed into the single-vector models, leveraging their native multi-image input support; and (III) \textbf{Late-chunking-layout}, which performs semantic chunking on the last-layer tokens based on explicit layout parsing boundaries. The layout-aware category is further split into two variants: \textit{late-chunking-layout-type}, which merges tokens of the same content type (\textit{e.g.,} text blocks, tables), and \textit{late-chunking-layout-subimg}, which aggregates tokens within each sub-image. While these baselines rely on explicit, parsing-heavy segmentations, \ourmethod offers a more streamlined, implicit chunking paradigm that achieves context preservation without requiring rigid layout constraints.

\textbf{Implementation Details.} 
To ensure a fair comparison, our entire evaluation pipeline is built upon MMEB framework \cite{meng2025vlm2vec}. For baselines requiring explicit document structure, we utilize MinerU2.5 \cite{niu2025mineru25} as the layout parsing engine to extract sub-images, bounding boxes, and content types. All experiments are conducted on a computing cluster equipped with NVIDIA A100 (80GB) GPUs.




\subsection{Experimental Analysis}

\subsubsection{Main Result}

As illustrated in Figure \ref{fig:main_result_visualization},  \ourmethod significantly enhances the retrieval efficacy of existing single-vector models by introducing an efficient late interaction mechanism. 
For example, \ourmethod achieves a substantial performance leap across all tested backbones, notably elevating the overall average score of VLM2Vec-7B from 32.74 to 51.32, a relative improvement of nearly 57\%. 
This gain is attributed to the contextualized multi-vector representation, which preserves fine-grained local semantic features that are typically collapsed in single-vector compression. 
Second, implicit semantic chunking within the embedding space proves superior to explicit layout-based partitioning that relies on external parsing engines. 
On ViDoRe-V1 benchmark with VLM2Vec-7B, \ourmethod (55.89) significantly outperforms the layout-type (21.63) and layout-subimg (27.12) baselines. 
This suggests that explicit parsing often introduces noise or arbitrarily severs semantic continuity, whereas our hierarchical clustering adaptively identifies optimal semantic boundaries, ensuring higher representational integrity. 
Finally, \ourmethod exhibits remarkable robustness and model-agnostic generalizability across diverse architectures. 
For instance, while LamRA-Ret-7B suffers a drastic performance drop when processing multi-image inputs (dropping to 14.28), \ourmethod consistently recovers and boosts its overall performance to 34.21. 

\subsubsection{Scaling Analysis}

As depicted in Figure~\ref{fig:chunk_size_scaling}, increasing the number of chunks consistently improves retrieval performance, but the gains exhibit diminishing returns, suggesting a highly efficient performance-to-storage trade-off at moderate chunk sizes.
For instance, on UniME-V2-7B, increasing the chunk size from 10 to 40 yields a significant 7.0-point gain in overall score (from 50.67 to 57.67), whereas doubling it again to 80 provides a smaller 2.03-point improvement.
This pattern indicates while a larger number of chunks better preserves fine-grained document details, a moderate number (\textit{e.g.,} 40) is sufficient to capture the most salient semantic regions, with further increases yielding marginal benefits.
Besides, using an overly small number of chunks can be detrimental, leading to performance degradation even below that of the original single-vector baseline.
Specifically, when using only 5 chunks, VLM2Vec-2B's overall score drops to 21.96, a significant 9.22-point decrease from its base performance of 31.18.
This is likely because extreme compression forces the merging of semantically disparate regions, leading to over-generalized chunk embeddings that obscure critical information more than the original global embedding.

\subsubsection{Ablation \& Hyperparameter Study}

\paragraph{\textbf{Effect of weighting factor.}} 

The weighting factor $\omega$ plays a crucial role in balancing spatial contiguity and semantic coherence, with a moderate spatial constraint (typically around $\omega=0.2$) yielding the best retrieval performance. 
As illustrated in Figure~\ref{fig:hyperparameters_comparison} (Left), for UniME-V2-2B with $K=80$, introducing $\omega=0.2$ increases the overall score from 49.64 to 51.40; similarly, VLM2Vec-2B at $K=40$ achieves its peak performance of 34.80 at $\omega=0.2$, whereas a further increase to $\omega=0.6$ results in a decline to 34.37. 
This suggests that while 2D positional priors help maintain the structural integrity of chunks, an excessive weight may overshadow the semantic features. 

\paragraph{\textbf{Effect of chunking location.}} 

Performing \ourmethod at the output of the LLM backbone is significantly more effective than at the vision encoder stage, as the former leverages the high-level semantic context essential for precise document retrieval. 
As illustrated in Figure~\ref{fig:hyperparameters_comparison} (Middle), for UniME-V2-7B, applying chunking after the LLM backbone achieves a peak overall score of 55.99, while performing it after vision encoder leads to a sharp decline to 40.66, even underperforming the single-vector base (48.41). 
Such results indicate LLM's transformer layers are crucial for injecting global context into patch embeddings, and premature chunking at the vision encoder likely discards fine-grained information before its semantic relationship with the rest of document is fully established.

\paragraph{\textbf{Effect of clustering method.}} 

Hierarchical agglomerative clustering (HAC) proves more effective than k-means for generating contextual chunks, consistently yielding superior retrieval performance. As shown in Figure~\ref{fig:hyperparameters_comparison} (Right), this superiority is evident on UniME-V2-7B, where HAC achieves an overall score of 55.99, outperforming k-means (50.88) by a 5.11-point margin. This performance discrepancy likely arises because HAC's ability to form clusters of arbitrary shapes and sizes is better suited for capturing the non-uniform structure of visual document elements, unlike k-means which produces more uniformly sized, spherical clusters.

\vspace{-2mm}

%% file: sections/conclusion.tex
In this paper, we introduced \ourmethod{}, a novel paradigm for efficient VDR. 
By adapting the concept of late chunking to the multimodal domain, we effectively addressed the storage bottleneck inherent in multi-vector architectures. 
Specifically, \ourmethod{} leverages HAC combined with a spatial position prior to generate contextualized representations. 
Extensive experiment across 24 VDR datasets confirm that our method significantly outperforms compression baselines, achieving a superior trade-off between retrieval quality and computational overhead. 
We believe \ourmethod{} can establishe a solid foundation for deploying advanced multimodal retrieval models in large-scale, real-world environments.

%% file: colchunk.bbl

\begin{thebibliography}{46}


\ifx \showCODEN    \undefined \def \showCODEN     #1{\unskip}     \fi
\ifx \showISBNx    \undefined \def \showISBNx     #1{\unskip}     \fi
\ifx \showISBNxiii \undefined \def \showISBNxiii  #1{\unskip}     \fi
\ifx \showISSN     \undefined \def \showISSN      #1{\unskip}     \fi
\ifx \showLCCN     \undefined \def \showLCCN      #1{\unskip}     \fi
\ifx \shownote     \undefined \def \shownote      #1{#1}          \fi
\ifx \showarticletitle \undefined \def \showarticletitle #1{#1}   \fi
\ifx \showURL      \undefined \def \showURL       {\relax}        \fi
\providecommand\bibfield[2]{#2}
\providecommand\bibinfo[2]{#2}
\providecommand\natexlab[1]{#1}
\providecommand\showeprint[2][]{arXiv:#2}

\bibitem[Cha et~al\mbox{.}(2026)]%
        {cha2026reinpool}
\bibfield{author}{\bibinfo{person}{Sungguk Cha}, \bibinfo{person}{DongWook Kim}, \bibinfo{person}{Mintae Kim}, \bibinfo{person}{Youngsub Han}, \bibinfo{person}{Byoung-Ki Jeon}, {and} \bibinfo{person}{Sangyeob Lee}.} \bibinfo{year}{2026}\natexlab{}.
\newblock \showarticletitle{ReinPool: Reinforcement Learning Pooling Multi-Vector Embeddings for Retrieval System}.
\newblock \bibinfo{journal}{\emph{arXiv preprint arXiv:2601.07125}} (\bibinfo{year}{2026}).
\newblock


\bibitem[Clavi{\'e} et~al\mbox{.}(2024)]%
        {clavie2024reducing}
\bibfield{author}{\bibinfo{person}{Benjamin Clavi{\'e}}, \bibinfo{person}{Antoine Chaffin}, {and} \bibinfo{person}{Griffin Adams}.} \bibinfo{year}{2024}\natexlab{}.
\newblock \showarticletitle{Reducing the footprint of multi-vector retrieval with minimal performance impact via token pooling}.
\newblock \bibinfo{journal}{\emph{arXiv preprint arXiv:2409.14683}} (\bibinfo{year}{2024}).
\newblock


\bibitem[Conti et~al\mbox{.}(2025)]%
        {conti2025context}
\bibfield{author}{\bibinfo{person}{Max Conti}, \bibinfo{person}{Manuel Faysse}, \bibinfo{person}{Gautier Viaud}, \bibinfo{person}{Antoine Bosselut}, \bibinfo{person}{C{\'e}line Hudelot}, {and} \bibinfo{person}{Pierre Colombo}.} \bibinfo{year}{2025}\natexlab{}.
\newblock \showarticletitle{Context is Gold to find the Gold Passage: Evaluating and Training Contextual Document Embeddings}.
\newblock \bibinfo{journal}{\emph{arXiv preprint arXiv:2505.24782}} (\bibinfo{year}{2025}).
\newblock


\bibitem[Ding et~al\mbox{.}(2025)]%
        {ding2025survey}
\bibfield{author}{\bibinfo{person}{Yihao Ding}, \bibinfo{person}{Siwen Luo}, \bibinfo{person}{Yue Dai}, \bibinfo{person}{Yanbei Jiang}, \bibinfo{person}{Zechuan Li}, \bibinfo{person}{Geoffrey Martin}, {and} \bibinfo{person}{Yifan Peng}.} \bibinfo{year}{2025}\natexlab{}.
\newblock \showarticletitle{A Survey on MLLM-based Visually Rich Document Understanding: Methods, Challenges, and Emerging Trends}.
\newblock \bibinfo{journal}{\emph{arXiv preprint arXiv:2507.09861}} (\bibinfo{year}{2025}).
\newblock


\bibitem[Faysse et~al\mbox{.}(2024)]%
        {faysse2024colpali}
\bibfield{author}{\bibinfo{person}{Manuel Faysse}, \bibinfo{person}{Hugues Sibille}, \bibinfo{person}{Tony Wu}, \bibinfo{person}{Bilel Omrani}, \bibinfo{person}{Gautier Viaud}, \bibinfo{person}{C{\'e}line Hudelot}, {and} \bibinfo{person}{Pierre Colombo}.} \bibinfo{year}{2024}\natexlab{}.
\newblock \showarticletitle{Colpali: Efficient document retrieval with vision language models}.
\newblock \bibinfo{journal}{\emph{arXiv preprint arXiv:2407.01449}} (\bibinfo{year}{2024}).
\newblock


\bibitem[Gao et~al\mbox{.}(2025)]%
        {gao2025scaling}
\bibfield{author}{\bibinfo{person}{Sensen Gao}, \bibinfo{person}{Shanshan Zhao}, \bibinfo{person}{Xu Jiang}, \bibinfo{person}{Lunhao Duan}, \bibinfo{person}{Yong~Xien Chng}, \bibinfo{person}{Qing-Guo Chen}, \bibinfo{person}{Weihua Luo}, \bibinfo{person}{Kaifu Zhang}, \bibinfo{person}{Jia-Wang Bian}, {and} \bibinfo{person}{Mingming Gong}.} \bibinfo{year}{2025}\natexlab{}.
\newblock \showarticletitle{Scaling Beyond Context: A Survey of Multimodal Retrieval-Augmented Generation for Document Understanding}.
\newblock \bibinfo{journal}{\emph{arXiv preprint arXiv:2510.15253}} (\bibinfo{year}{2025}).
\newblock


\bibitem[Gu et~al\mbox{.}(2025)]%
        {gu2025unimev2}
\bibfield{author}{\bibinfo{person}{Tiancheng Gu}, \bibinfo{person}{Kaicheng Yang}, \bibinfo{person}{Kaichen Zhang}, \bibinfo{person}{Xiang An}, \bibinfo{person}{Ziyong Feng}, \bibinfo{person}{Yueyi Zhang}, \bibinfo{person}{Weidong Cai}, \bibinfo{person}{Jiankang Deng}, {and} \bibinfo{person}{Lidong Bing}.} \bibinfo{year}{2025}\natexlab{}.
\newblock \bibinfo{title}{UniME-V2: MLLM-as-a-Judge for Universal Multimodal Embedding Learning}.
\newblock
\showeprint[arxiv]{2510.13515}~[cs.CV]
\urldef\tempurl%
\url{https://arxiv.org/abs/2510.13515}
\showURL{%
\tempurl}


\bibitem[G{\"u}nther et~al\mbox{.}(2024)]%
        {gunther2024late}
\bibfield{author}{\bibinfo{person}{Michael G{\"u}nther}, \bibinfo{person}{Isabelle Mohr}, \bibinfo{person}{Daniel~James Williams}, \bibinfo{person}{Bo Wang}, {and} \bibinfo{person}{Han Xiao}.} \bibinfo{year}{2024}\natexlab{}.
\newblock \showarticletitle{Late chunking: contextual chunk embeddings using long-context embedding models}.
\newblock \bibinfo{journal}{\emph{arXiv preprint arXiv:2409.04701}} (\bibinfo{year}{2024}).
\newblock


\bibitem[G{\"u}nther et~al\mbox{.}(2025)]%
        {gunther2025jina}
\bibfield{author}{\bibinfo{person}{Michael G{\"u}nther}, \bibinfo{person}{Saba Sturua}, \bibinfo{person}{Mohammad~Kalim Akram}, \bibinfo{person}{Isabelle Mohr}, \bibinfo{person}{Andrei Ungureanu}, \bibinfo{person}{Bo Wang}, \bibinfo{person}{Sedigheh Eslami}, \bibinfo{person}{Scott Martens}, \bibinfo{person}{Maximilian Werk}, \bibinfo{person}{Nan Wang}, {et~al\mbox{.}}} \bibinfo{year}{2025}\natexlab{}.
\newblock \showarticletitle{jina-embeddings-v4: Universal Embeddings for Multimodal Multilingual Retrieval}.
\newblock \bibinfo{journal}{\emph{arXiv preprint arXiv:2506.18902}} (\bibinfo{year}{2025}).
\newblock


\bibitem[He et~al\mbox{.}(2025)]%
        {he2025token}
\bibfield{author}{\bibinfo{person}{Shanxiu He}, \bibinfo{person}{Mutasem Al-Darabsah}, \bibinfo{person}{Suraj Nair}, \bibinfo{person}{Jonathan May}, \bibinfo{person}{Tarun Agarwal}, \bibinfo{person}{Tao Yang}, {and} \bibinfo{person}{Choon~Hui Teo}.} \bibinfo{year}{2025}\natexlab{}.
\newblock \showarticletitle{Token pruning optimization for efficient multi-vector dense retrieval}. In \bibinfo{booktitle}{\emph{European Conference on Information Retrieval}}. Springer, \bibinfo{pages}{101--115}.
\newblock


\bibitem[Huo et~al\mbox{.}(2026)]%
        {huo2026causalembed}
\bibfield{author}{\bibinfo{person}{Jiahao Huo}, \bibinfo{person}{Yu Huang}, \bibinfo{person}{Yibo Yan}, \bibinfo{person}{Ye Pan}, \bibinfo{person}{Yi Cao}, \bibinfo{person}{Mingdong Ou}, \bibinfo{person}{Philip~S Yu}, {and} \bibinfo{person}{Xuming Hu}.} \bibinfo{year}{2026}\natexlab{}.
\newblock \showarticletitle{CausalEmbed: Auto-Regressive Multi-Vector Generation in Latent Space for Visual Document Embedding}.
\newblock \bibinfo{journal}{\emph{arXiv preprint arXiv:2601.21262}} (\bibinfo{year}{2026}).
\newblock


\bibitem[Jiang et~al\mbox{.}(2024)]%
        {jiang2024vlm2vec}
\bibfield{author}{\bibinfo{person}{Ziyan Jiang}, \bibinfo{person}{Rui Meng}, \bibinfo{person}{Xinyi Yang}, \bibinfo{person}{Semih Yavuz}, \bibinfo{person}{Yingbo Zhou}, {and} \bibinfo{person}{Wenhu Chen}.} \bibinfo{year}{2024}\natexlab{}.
\newblock \showarticletitle{Vlm2vec: Training vision-language models for massive multimodal embedding tasks}.
\newblock \bibinfo{journal}{\emph{arXiv preprint arXiv:2410.05160}} (\bibinfo{year}{2024}).
\newblock


\bibitem[Khattab and Zaharia(2020)]%
        {khattab2020colbert}
\bibfield{author}{\bibinfo{person}{Omar Khattab} {and} \bibinfo{person}{Matei Zaharia}.} \bibinfo{year}{2020}\natexlab{}.
\newblock \showarticletitle{Colbert: Efficient and effective passage search via contextualized late interaction over bert}. In \bibinfo{booktitle}{\emph{Proceedings of the 43rd International ACM SIGIR conference on research and development in Information Retrieval}}. \bibinfo{pages}{39--48}.
\newblock


\bibitem[Lassance et~al\mbox{.}(2022)]%
        {lassance2022learned}
\bibfield{author}{\bibinfo{person}{Carlos Lassance}, \bibinfo{person}{Maroua Maachou}, \bibinfo{person}{Joohee Park}, {and} \bibinfo{person}{St{\'e}phane Clinchant}.} \bibinfo{year}{2022}\natexlab{}.
\newblock \showarticletitle{Learned token pruning in contextualized late interaction over BERT (ColBERT)}. In \bibinfo{booktitle}{\emph{Proceedings of the 45th International ACM SIGIR Conference on Research and Development in Information Retrieval}}. \bibinfo{pages}{2232--2236}.
\newblock


\bibitem[Lee et~al\mbox{.}(2023)]%
        {lee2023rethinking}
\bibfield{author}{\bibinfo{person}{Jinhyuk Lee}, \bibinfo{person}{Zhuyun Dai}, \bibinfo{person}{Sai Meher~Karthik Duddu}, \bibinfo{person}{Tao Lei}, \bibinfo{person}{Iftekhar Naim}, \bibinfo{person}{Ming-Wei Chang}, {and} \bibinfo{person}{Vincent Zhao}.} \bibinfo{year}{2023}\natexlab{}.
\newblock \showarticletitle{Rethinking the role of token retrieval in multi-vector retrieval}.
\newblock \bibinfo{journal}{\emph{Advances in Neural Information Processing Systems}}  \bibinfo{volume}{36} (\bibinfo{year}{2023}), \bibinfo{pages}{15384--15405}.
\newblock


\bibitem[Liu and Mao(2023)]%
        {liu2023understanding}
\bibfield{author}{\bibinfo{person}{Qi Liu} {and} \bibinfo{person}{Jiaxin Mao}.} \bibinfo{year}{2023}\natexlab{}.
\newblock \showarticletitle{Understanding the Multi-vector Dense Retrieval Models}. In \bibinfo{booktitle}{\emph{Proceedings of the 32nd ACM International Conference on Information and Knowledge Management}}. \bibinfo{pages}{4110--4114}.
\newblock


\bibitem[Liu et~al\mbox{.}(2025b)]%
        {liu2025lamra}
\bibfield{author}{\bibinfo{person}{Yikun Liu}, \bibinfo{person}{Yajie Zhang}, \bibinfo{person}{Jiayin Cai}, \bibinfo{person}{Xiaolong Jiang}, \bibinfo{person}{Yao Hu}, \bibinfo{person}{Jiangchao Yao}, \bibinfo{person}{Yanfeng Wang}, {and} \bibinfo{person}{Weidi Xie}.} \bibinfo{year}{2025}\natexlab{b}.
\newblock \showarticletitle{Lamra: Large multimodal model as your advanced retrieval assistant}. In \bibinfo{booktitle}{\emph{Proceedings of the Computer Vision and Pattern Recognition Conference}}. \bibinfo{pages}{4015--4025}.
\newblock


\bibitem[Liu et~al\mbox{.}(2025a)]%
        {liu2025any}
\bibfield{author}{\bibinfo{person}{Zheng Liu}, \bibinfo{person}{Ze Liu}, \bibinfo{person}{Zhengyang Liang}, \bibinfo{person}{Junjie Zhou}, \bibinfo{person}{Shitao Xiao}, \bibinfo{person}{Chao Gao}, \bibinfo{person}{Chen~Jason Zhang}, {and} \bibinfo{person}{Defu Lian}.} \bibinfo{year}{2025}\natexlab{a}.
\newblock \showarticletitle{Any information is just worth one single screenshot: Unifying search with visualized information retrieval}. In \bibinfo{booktitle}{\emph{Proceedings of the 63rd Annual Meeting of the Association for Computational Linguistics (Volume 1: Long Papers)}}. \bibinfo{pages}{19238--19261}.
\newblock


\bibitem[Ma et~al\mbox{.}(2024a)]%
        {ma2024dse}
\bibfield{author}{\bibinfo{person}{Xueguang Ma}, \bibinfo{person}{Sheng-Chieh Lin}, \bibinfo{person}{Minghan Li}, \bibinfo{person}{Wenhu Chen}, {and} \bibinfo{person}{Jimmy Lin}.} \bibinfo{year}{2024}\natexlab{a}.
\newblock \showarticletitle{Unifying Multimodal Retrieval via Document Screenshot Embedding}. In \bibinfo{booktitle}{\emph{Proceedings of the 2024 Conference on Empirical Methods in Natural Language Processing}}, \bibfield{editor}{\bibinfo{person}{Yaser Al-Onaizan}, \bibinfo{person}{Mohit Bansal}, {and} \bibinfo{person}{Yun-Nung Chen}} (Eds.). \bibinfo{publisher}{Association for Computational Linguistics}, \bibinfo{address}{Miami, Florida, USA}, \bibinfo{pages}{6492--6505}.
\newblock
\href{https://doi.org/10.18653/v1/2024.emnlp-main.373}{doi:\nolinkurl{10.18653/v1/2024.emnlp-main.373}}


\bibitem[Ma et~al\mbox{.}(2025)]%
        {ma2025towards}
\bibfield{author}{\bibinfo{person}{Yubo Ma}, \bibinfo{person}{Jinsong Li}, \bibinfo{person}{Yuhang Zang}, \bibinfo{person}{Xiaobao Wu}, \bibinfo{person}{Xiaoyi Dong}, \bibinfo{person}{Pan Zhang}, \bibinfo{person}{Yuhang Cao}, \bibinfo{person}{Haodong Duan}, \bibinfo{person}{Jiaqi Wang}, \bibinfo{person}{Yixin Cao}, {et~al\mbox{.}}} \bibinfo{year}{2025}\natexlab{}.
\newblock \showarticletitle{Towards Storage-Efficient Visual Document Retrieval: An Empirical Study on Reducing Patch-Level Embeddings}.
\newblock \bibinfo{journal}{\emph{arXiv preprint arXiv:2506.04997}} (\bibinfo{year}{2025}).
\newblock


\bibitem[Ma et~al\mbox{.}(2024b)]%
        {ma2024mmlongbench}
\bibfield{author}{\bibinfo{person}{Yubo Ma}, \bibinfo{person}{Yuhang Zang}, \bibinfo{person}{Liangyu Chen}, \bibinfo{person}{Meiqi Chen}, \bibinfo{person}{Yizhu Jiao}, \bibinfo{person}{Xinze Li}, \bibinfo{person}{Xinyuan Lu}, \bibinfo{person}{Ziyu Liu}, \bibinfo{person}{Yan Ma}, \bibinfo{person}{Xiaoyi Dong}, {et~al\mbox{.}}} \bibinfo{year}{2024}\natexlab{b}.
\newblock \showarticletitle{Mmlongbench-doc: Benchmarking long-context document understanding with visualizations}.
\newblock \bibinfo{journal}{\emph{Advances in Neural Information Processing Systems}}  \bibinfo{volume}{37} (\bibinfo{year}{2024}), \bibinfo{pages}{95963--96010}.
\newblock


\bibitem[MacAvaney et~al\mbox{.}(2025)]%
        {macavaney2025efficient}
\bibfield{author}{\bibinfo{person}{Sean MacAvaney}, \bibinfo{person}{Antonio Mallia}, {and} \bibinfo{person}{Nicola Tonellotto}.} \bibinfo{year}{2025}\natexlab{}.
\newblock \showarticletitle{Efficient Constant-Space Multi-vector Retrieval}. In \bibinfo{booktitle}{\emph{European Conference on Information Retrieval}}. Springer, \bibinfo{pages}{237--245}.
\newblock


\bibitem[Mac{\'e} et~al\mbox{.}(2025)]%
        {mace2025vidorev2}
\bibfield{author}{\bibinfo{person}{Quentin Mac{\'e}}, \bibinfo{person}{Ant{\'o}nio Loison}, {and} \bibinfo{person}{Manuel Faysse}.} \bibinfo{year}{2025}\natexlab{}.
\newblock \showarticletitle{ViDoRe Benchmark V2: Raising the Bar for Visual Retrieval}.
\newblock \bibinfo{journal}{\emph{arXiv preprint arXiv:2505.17166}} (\bibinfo{year}{2025}).
\newblock


\bibitem[Meng et~al\mbox{.}(2025)]%
        {meng2025vlm2vec}
\bibfield{author}{\bibinfo{person}{Rui Meng}, \bibinfo{person}{Ziyan Jiang}, \bibinfo{person}{Ye Liu}, \bibinfo{person}{Mingyi Su}, \bibinfo{person}{Xinyi Yang}, \bibinfo{person}{Yuepeng Fu}, \bibinfo{person}{Can Qin}, \bibinfo{person}{Zeyuan Chen}, \bibinfo{person}{Ran Xu}, \bibinfo{person}{Caiming Xiong}, {et~al\mbox{.}}} \bibinfo{year}{2025}\natexlab{}.
\newblock \showarticletitle{Vlm2vec-v2: Advancing multimodal embedding for videos, images, and visual documents}.
\newblock \bibinfo{journal}{\emph{arXiv preprint arXiv:2507.04590}} (\bibinfo{year}{2025}).
\newblock


\bibitem[Merola and Singh(2025)]%
        {merola2025reconstructing}
\bibfield{author}{\bibinfo{person}{Carlo Merola} {and} \bibinfo{person}{Jaspinder Singh}.} \bibinfo{year}{2025}\natexlab{}.
\newblock \showarticletitle{Reconstructing context: Evaluating advanced chunking strategies for retrieval-augmented generation}. In \bibinfo{booktitle}{\emph{International Workshop on Knowledge-Enhanced Information Retrieval}}. Springer, \bibinfo{pages}{3--18}.
\newblock


\bibitem[Moreira et~al\mbox{.}(2026)]%
        {moreira2026nemotron}
\bibfield{author}{\bibinfo{person}{Gabriel de Souza~P Moreira}, \bibinfo{person}{Ronay Ak}, \bibinfo{person}{Mengyao Xu}, \bibinfo{person}{Oliver Holworthy}, \bibinfo{person}{Benedikt Schifferer}, \bibinfo{person}{Zhiding Yu}, \bibinfo{person}{Yauhen Babakhin}, \bibinfo{person}{Radek Osmulski}, \bibinfo{person}{Jiarui Cai}, \bibinfo{person}{Ryan Chesler}, {et~al\mbox{.}}} \bibinfo{year}{2026}\natexlab{}.
\newblock \showarticletitle{Nemotron ColEmbed V2: Top-Performing Late Interaction embedding models for Visual Document Retrieval}.
\newblock \bibinfo{journal}{\emph{arXiv preprint arXiv:2602.03992}} (\bibinfo{year}{2026}).
\newblock


\bibitem[Most et~al\mbox{.}(2025)]%
        {most2025lost}
\bibfield{author}{\bibinfo{person}{Alexander Most}, \bibinfo{person}{Joseph Winjum}, \bibinfo{person}{Manish Bhattarai}, \bibinfo{person}{Shawn Jones}, \bibinfo{person}{Nishath~Rajiv Ranasinghe}, \bibinfo{person}{Ayan Biswas}, {and} \bibinfo{person}{Dan O'Malley}.} \bibinfo{year}{2025}\natexlab{}.
\newblock \showarticletitle{Lost in ocr translation? vision-based approaches to robust document retrieval}. In \bibinfo{booktitle}{\emph{Proceedings of the 2025 ACM Symposium on Document Engineering}}. \bibinfo{pages}{1--10}.
\newblock


\bibitem[Niu et~al\mbox{.}(2025)]%
        {niu2025mineru25}
\bibfield{author}{\bibinfo{person}{Junbo Niu}, \bibinfo{person}{Zheng Liu}, \bibinfo{person}{Zhuangcheng Gu}, \bibinfo{person}{Bin Wang}, \bibinfo{person}{Linke Ouyang}, \bibinfo{person}{Zhiyuan Zhao}, \bibinfo{person}{Tao Chu}, \bibinfo{person}{Tianyao He}, \bibinfo{person}{Fan Wu}, \bibinfo{person}{Qintong Zhang}, {et~al\mbox{.}}} \bibinfo{year}{2025}\natexlab{}.
\newblock \showarticletitle{Mineru2. 5: A decoupled vision-language model for efficient high-resolution document parsing}.
\newblock \bibinfo{journal}{\emph{arXiv preprint arXiv:2509.22186}} (\bibinfo{year}{2025}).
\newblock


\bibitem[Park et~al\mbox{.}(2025)]%
        {park2025scv}
\bibfield{author}{\bibinfo{person}{Cheoneum Park}, \bibinfo{person}{Seohyeong Jeong}, \bibinfo{person}{Minsang Kim}, \bibinfo{person}{KyungTae Lim}, {and} \bibinfo{person}{Yong-Hun Lee}.} \bibinfo{year}{2025}\natexlab{}.
\newblock \showarticletitle{SCV: Light and Effective Multi-Vector Retrieval with Sequence Compressive Vectors}. In \bibinfo{booktitle}{\emph{Proceedings of the 31st International Conference on Computational Linguistics: Industry Track}}. \bibinfo{pages}{760--770}.
\newblock


\bibitem[Qian et~al\mbox{.}(2022)]%
        {qian2022multi}
\bibfield{author}{\bibinfo{person}{Yujie Qian}, \bibinfo{person}{Jinhyuk Lee}, \bibinfo{person}{Sai Meher~Karthik Duddu}, \bibinfo{person}{Zhuyun Dai}, \bibinfo{person}{Siddhartha Brahma}, \bibinfo{person}{Iftekhar Naim}, \bibinfo{person}{Tao Lei}, {and} \bibinfo{person}{Vincent~Y Zhao}.} \bibinfo{year}{2022}\natexlab{}.
\newblock \showarticletitle{Multi-vector retrieval as sparse alignment}.
\newblock \bibinfo{journal}{\emph{arXiv preprint arXiv:2211.01267}} (\bibinfo{year}{2022}).
\newblock


\bibitem[Santhanam et~al\mbox{.}(2022)]%
        {santhanam2022colbertv2}
\bibfield{author}{\bibinfo{person}{Keshav Santhanam}, \bibinfo{person}{Omar Khattab}, \bibinfo{person}{Jon Saad-Falcon}, \bibinfo{person}{Christopher Potts}, {and} \bibinfo{person}{Matei Zaharia}.} \bibinfo{year}{2022}\natexlab{}.
\newblock \showarticletitle{Colbertv2: Effective and efficient retrieval via lightweight late interaction}. In \bibinfo{booktitle}{\emph{Proceedings of the 2022 Conference of the North American Chapter of the Association for Computational Linguistics: Human Language Technologies}}. \bibinfo{pages}{3715--3734}.
\newblock


\bibitem[Scheerer et~al\mbox{.}(2025)]%
        {scheerer2025warp}
\bibfield{author}{\bibinfo{person}{Jan~Luca Scheerer}, \bibinfo{person}{Matei Zaharia}, \bibinfo{person}{Christopher Potts}, \bibinfo{person}{Gustavo Alonso}, {and} \bibinfo{person}{Omar Khattab}.} \bibinfo{year}{2025}\natexlab{}.
\newblock \showarticletitle{WARP: An efficient engine for multi-vector retrieval}. In \bibinfo{booktitle}{\emph{Proceedings of the 48th international ACM SIGIR conference on research and development in information retrieval}}. \bibinfo{pages}{2504--2512}.
\newblock


\bibitem[Shrestha et~al\mbox{.}(2024)]%
        {shrestha2024espn}
\bibfield{author}{\bibinfo{person}{Susav Shrestha}, \bibinfo{person}{Narasimha Reddy}, {and} \bibinfo{person}{Zongwang Li}.} \bibinfo{year}{2024}\natexlab{}.
\newblock \showarticletitle{Espn: Memory-efficient multi-vector information retrieval}. In \bibinfo{booktitle}{\emph{Proceedings of the 2024 ACM SIGPLAN International Symposium on Memory Management}}. \bibinfo{pages}{95--107}.
\newblock


\bibitem[Teiletche et~al\mbox{.}(2025)]%
        {teiletche2025modernvbert}
\bibfield{author}{\bibinfo{person}{Paul Teiletche}, \bibinfo{person}{Quentin Mac{\'e}}, \bibinfo{person}{Max Conti}, \bibinfo{person}{Antonio Loison}, \bibinfo{person}{Gautier Viaud}, \bibinfo{person}{Pierre Colombo}, {and} \bibinfo{person}{Manuel Faysse}.} \bibinfo{year}{2025}\natexlab{}.
\newblock \showarticletitle{ModernVBERT: Towards Smaller Visual Document Retrievers}.
\newblock \bibinfo{journal}{\emph{arXiv preprint arXiv:2510.01149}} (\bibinfo{year}{2025}).
\newblock


\bibitem[Tishby et~al\mbox{.}(2000)]%
        {tishby2000information}
\bibfield{author}{\bibinfo{person}{Naftali Tishby}, \bibinfo{person}{Fernando~C Pereira}, {and} \bibinfo{person}{William Bialek}.} \bibinfo{year}{2000}\natexlab{}.
\newblock \showarticletitle{The information bottleneck method}.
\newblock \bibinfo{journal}{\emph{arXiv preprint physics/0004057}} (\bibinfo{year}{2000}).
\newblock


\bibitem[Veneroso et~al\mbox{.}(2025)]%
        {veneroso2025crisp}
\bibfield{author}{\bibinfo{person}{Jo{\~a}o Veneroso}, \bibinfo{person}{Rajesh Jayaram}, \bibinfo{person}{Jinmeng Rao}, \bibinfo{person}{Gustavo~Hern{\'a}ndez {\'A}brego}, \bibinfo{person}{Majid Hadian}, {and} \bibinfo{person}{Daniel Cer}.} \bibinfo{year}{2025}\natexlab{}.
\newblock \showarticletitle{CRISP: Clustering Multi-Vector Representations for Denoising and Pruning}.
\newblock \bibinfo{journal}{\emph{arXiv preprint arXiv:2505.11471}} (\bibinfo{year}{2025}).
\newblock


\bibitem[Wang et~al\mbox{.}(2025b)]%
        {wang2025jinarerankerv3}
\bibfield{author}{\bibinfo{person}{Feng Wang}, \bibinfo{person}{Yuqing Li}, {and} \bibinfo{person}{Han Xiao}.} \bibinfo{year}{2025}\natexlab{b}.
\newblock \showarticletitle{Jina-reranker-v3: Last but Not Late Interaction for Listwise Document Reranking}.
\newblock \bibinfo{journal}{\emph{arXiv preprint arXiv:2509.25085}} (\bibinfo{year}{2025}).
\newblock


\bibitem[Wang et~al\mbox{.}(2025a)]%
        {wang2025vidorag}
\bibfield{author}{\bibinfo{person}{Qiuchen Wang}, \bibinfo{person}{Ruixue Ding}, \bibinfo{person}{Zehui Chen}, \bibinfo{person}{Weiqi Wu}, \bibinfo{person}{Shihang Wang}, \bibinfo{person}{Pengjun Xie}, {and} \bibinfo{person}{Feng Zhao}.} \bibinfo{year}{2025}\natexlab{a}.
\newblock \showarticletitle{Vidorag: Visual document retrieval-augmented generation via dynamic iterative reasoning agents}.
\newblock \bibinfo{journal}{\emph{arXiv preprint arXiv:2502.18017}} (\bibinfo{year}{2025}).
\newblock


\bibitem[Xiao et~al\mbox{.}(2025)]%
        {xiao2025metaembed}
\bibfield{author}{\bibinfo{person}{Zilin Xiao}, \bibinfo{person}{Qi Ma}, \bibinfo{person}{Mengting Gu}, \bibinfo{person}{Chun-cheng~Jason Chen}, \bibinfo{person}{Xintao Chen}, \bibinfo{person}{Vicente Ordonez}, {and} \bibinfo{person}{Vijai Mohan}.} \bibinfo{year}{2025}\natexlab{}.
\newblock \showarticletitle{Metaembed: Scaling multimodal retrieval at test-time with flexible late interaction}.
\newblock \bibinfo{journal}{\emph{arXiv preprint arXiv:2509.18095}} (\bibinfo{year}{2025}).
\newblock


\bibitem[Yan et~al\mbox{.}(2026a)]%
        {yan2026unlocking}
\bibfield{author}{\bibinfo{person}{Yibo Yan}, \bibinfo{person}{Jiahao Huo}, \bibinfo{person}{Guanbo Feng}, \bibinfo{person}{Mingdong Ou}, \bibinfo{person}{Yi Cao}, \bibinfo{person}{Xin Zou}, \bibinfo{person}{Shuliang Liu}, \bibinfo{person}{Yuanhuiyi Lyu}, \bibinfo{person}{Yu Huang}, \bibinfo{person}{Jungang Li}, {et~al\mbox{.}}} \bibinfo{year}{2026}\natexlab{a}.
\newblock \showarticletitle{Unlocking Multimodal Document Intelligence: From Current Triumphs to Future Frontiers of Visual Document Retrieval}.
\newblock \bibinfo{journal}{\emph{arXiv preprint arXiv:2602.19961}} (\bibinfo{year}{2026}).
\newblock


\bibitem[Yan et~al\mbox{.}(2026b)]%
        {yan2026sculpting}
\bibfield{author}{\bibinfo{person}{Yibo Yan}, \bibinfo{person}{Mingdong Ou}, \bibinfo{person}{Yi Cao}, \bibinfo{person}{Xin Zou}, \bibinfo{person}{Jiahao Huo}, \bibinfo{person}{Shuliang Liu}, \bibinfo{person}{James Kwok}, {and} \bibinfo{person}{Xuming Hu}.} \bibinfo{year}{2026}\natexlab{b}.
\newblock \showarticletitle{Sculpting the Vector Space: Towards Efficient Multi-Vector Visual Document Retrieval via Prune-then-Merge Framework}.
\newblock \bibinfo{journal}{\emph{arXiv preprint arXiv:2602.19549}} (\bibinfo{year}{2026}).
\newblock


\bibitem[Yan et~al\mbox{.}(2026c)]%
        {yan2026beyond}
\bibfield{author}{\bibinfo{person}{Yibo Yan}, \bibinfo{person}{Mingdong Ou}, \bibinfo{person}{Yi Cao}, \bibinfo{person}{Xin Zou}, \bibinfo{person}{Shuliang Liu}, \bibinfo{person}{Jiahao Huo}, \bibinfo{person}{Yu Huang}, \bibinfo{person}{James Kwok}, {and} \bibinfo{person}{Xuming Hu}.} \bibinfo{year}{2026}\natexlab{c}.
\newblock \showarticletitle{Beyond the Grid: Layout-Informed Multi-Vector Retrieval with Parsed Visual Document Representations}.
\newblock \bibinfo{journal}{\emph{arXiv preprint arXiv:2603.01666}} (\bibinfo{year}{2026}).
\newblock


\bibitem[Yan et~al\mbox{.}(2025)]%
        {yan2025docpruner}
\bibfield{author}{\bibinfo{person}{Yibo Yan}, \bibinfo{person}{Guangwei Xu}, \bibinfo{person}{Xin Zou}, \bibinfo{person}{Shuliang Liu}, \bibinfo{person}{James Kwok}, {and} \bibinfo{person}{Xuming Hu}.} \bibinfo{year}{2025}\natexlab{}.
\newblock \showarticletitle{Docpruner: A storage-efficient framework for multi-vector visual document retrieval via adaptive patch-level embedding pruning}.
\newblock \bibinfo{journal}{\emph{arXiv preprint arXiv:2509.23883}} (\bibinfo{year}{2025}).
\newblock


\bibitem[Yu et~al\mbox{.}(2024)]%
        {yu2024visrag}
\bibfield{author}{\bibinfo{person}{Shi Yu}, \bibinfo{person}{Chaoyue Tang}, \bibinfo{person}{Bokai Xu}, \bibinfo{person}{Junbo Cui}, \bibinfo{person}{Junhao Ran}, \bibinfo{person}{Yukun Yan}, \bibinfo{person}{Zhenghao Liu}, \bibinfo{person}{Shuo Wang}, \bibinfo{person}{Xu Han}, \bibinfo{person}{Zhiyuan Liu}, {et~al\mbox{.}}} \bibinfo{year}{2024}\natexlab{}.
\newblock \showarticletitle{Visrag: Vision-based retrieval-augmented generation on multi-modality documents}.
\newblock \bibinfo{journal}{\emph{arXiv preprint arXiv:2410.10594}} (\bibinfo{year}{2024}).
\newblock


\bibitem[Zhang et~al\mbox{.}(2025)]%
        {zhang2025ocr}
\bibfield{author}{\bibinfo{person}{Junyuan Zhang}, \bibinfo{person}{Qintong Zhang}, \bibinfo{person}{Bin Wang}, \bibinfo{person}{Linke Ouyang}, \bibinfo{person}{Zichen Wen}, \bibinfo{person}{Ying Li}, \bibinfo{person}{Ka-Ho Chow}, \bibinfo{person}{Conghui He}, {and} \bibinfo{person}{Wentao Zhang}.} \bibinfo{year}{2025}\natexlab{}.
\newblock \showarticletitle{Ocr hinders rag: Evaluating the cascading impact of ocr on retrieval-augmented generation}. In \bibinfo{booktitle}{\emph{Proceedings of the IEEE/CVF International Conference on Computer Vision}}. \bibinfo{pages}{17443--17453}.
\newblock


\bibitem[Zhang et~al\mbox{.}(2024)]%
        {zhang2024gme}
\bibfield{author}{\bibinfo{person}{Xin Zhang}, \bibinfo{person}{Yanzhao Zhang}, \bibinfo{person}{Wen Xie}, \bibinfo{person}{Mingxin Li}, \bibinfo{person}{Ziqi Dai}, \bibinfo{person}{Dingkun Long}, \bibinfo{person}{Pengjun Xie}, \bibinfo{person}{Meishan Zhang}, \bibinfo{person}{Wenjie Li}, {and} \bibinfo{person}{Min Zhang}.} \bibinfo{year}{2024}\natexlab{}.
\newblock \showarticletitle{GME: Improving Universal Multimodal Retrieval by Multimodal LLMs}.
\newblock \bibinfo{journal}{\emph{arXiv preprint arXiv:2412.16855}} (\bibinfo{year}{2024}).
\newblock


\end{thebibliography}
